# SecurePose: Automated Face Blurring and Human Movement Kinematics Extraction from Videos Recorded in Clinical Settings

Rishabh Bajpai, and Bhooma Aravamuthan

## Abstract

Movement disorder diagnosis often relies on expert evaluation of patient videos, but sharing these videos poses privacy risks. Current methods for de-identifying videos, such as blurring faces, are often manual, inconsistent, or inaccurate. Furthermore, these methods can compromise objective kinematic analysis – a crucial component of diagnosis. To address these challenges, we developed SecurePose, an open-source software that simultaneously provides reliable de-identification and automated kinematic extraction from videos recorded in clinic settings using smartphones/tablets. SecurePose utilizes pose estimation (using OpenPose) to extract full body kinematics, track individuals, identify the patient, and then accurately blur faces in the videos. We validated SecurePose on gait videos recorded in outpatient clinic visits of 116 children with cerebral palsy, assessing both the accuracy of its de-identification compared to the ground truth (manual blurring) and the reliability of the intermediate steps of kinematics extraction. Results demonstrate that SecurePose outperformed six existing methods in automated face detection and achieved comparable accuracy to robust manual blurring, but in significantly less time (91.08% faster). Ten experienced researchers also confirmed SecurePose's usability via System Usability Scale scores. These findings validate SecurePose as a practical and effective tool for protecting patient privacy while enabling accurate kinematics extraction in clinical settings.

## INTRODUCTION

The potential for dramatically improving clinical care and accelerating medical research hinges on the broad sharing of medical data. However, realizing this potential is currently hampered by significant challenges in protecting patient privacy while simultaneously extracting clinically valuable information. A particularly rich, yet sensitive, data source is video recordings of clinical visits, crucial for diagnosing and tracking progression in numerous conditions, especially motor disorders.[1–4] These recordings are increasingly used for detailed, quantitative assessments unattainable through traditional methods, driven by advancements in pose estimation and kinematic analysis.[5] However, responsibly sharing this data requires robust anonymization techniques, and existing methods often fall short, presenting a critical barrier to progress.

Current regulations, such as the Health Insurance Portability and Accountability Act (HIPAA)[6] in the United States, the General Data Protection Regulation (GDPR)[7] in the European Union, and the Personal Data Protection Bill (PDPB)[8] in India, all emphasize the sensitive nature of facial images and the need for stringent protection of personally identifiable information. Consequently, any attempt to share video data must prioritize complete removal of identifying features, such as faces, before storage or distribution. While manual blurring is the current gold standard, it is a time-consuming, labor-intensive, and prone-to-error process, hindering the large-scale data sharing necessary for robust research.

A spectrum of automated face detection methods have been proposed, each with its strengths and weaknesses. Early approaches relied on feature-based techniques, such as the Viola-Jones algorithm[9] and Histogram of Oriented Gradients (HOG).[10] These methods are computationally efficient and relatively straightforward to implement, making them suitable for real-time applications. However, they can struggle with variations in pose, lighting, and occlusion, potentially leading to missed detections or inaccurate bounding boxes.[9,10]

More recently, image-based techniques leveraging convolutional neural networks (CNNs) have gained prominence. Algorithms like Faster R-CNN,[11] Single Shot MultiBox Detector (SSD),[12] MMOD,[13] S$^3$FD[14] and YOLO[15] offer significantly improved accuracy and robustness compared to traditional feature-based methods.[16,17] These algorithms excel at detecting faces in complex scenes and under challenging conditions.[16,17] However, they require substantial computational resources and large training datasets, potentially limiting their practicality in resource-constrained settings.[16,17] MTCNN[18] offers a good balance between performance and efficiency, becoming a popular choice for many applications. However, even these advanced algorithms can struggle with specific scenarios, particularly those involving low-resolution video, unusual angles, or obscured facial features.[11–17] Furthermore, many existing algorithms were designed for static images and may not maintain temporal consistency across video frames, resulting in flickering or inconsistent blurring .

Extracting body kinematics from video recordings, facilitated by tools like OpenPose,[19] is revolutionizing the assessment of motor disorders, moving beyond subjective evaluations to quantifiable, objective measures. However, the accuracy of this kinematic data extraction is directly impacted by the effectiveness of face detection algorithms; a wrongly blurred non-face region can introduce errors into the resulting analysis while a correctly blurred face region may confuse the kinematic data extraction algorithm to detect other body parts.[20] Therefore, it is crucial to perform kinematic data extraction before applying anonymization techniques, as anonymizing the raw video beforehand can compromise the accuracy of the measurements and introduce unwanted errors.

To advance research in motor disorders, a dedicated pipeline optimized for clinical video data is crucial, prioritizing both accurate anonymization and reliable kinematic analysis. This requires a system capable of reliably identifying and obscuring patient faces while preserving the integrity of extracted kinematic data, overcoming the limitations of existing methods.

## METHODS

### A. Dataset description

We obtained ethical approval from the Washington University in St. Louis Institutional Review Board (approval number 202102101). Informed consent was obtained from all participants (and caregivers for minors) prior to their inclusion in the study and use of their clinical data, including video recordings. Participants were recruited from the St. Louis Children's Hospital Cerebral Palsy Clinic between January 1, 2020, and November 4, 2021. A total of 180 individuals met the following inclusion criteria: age 10-20 years at the time of evaluation, diagnosis of cerebral palsy according to the 2006 consensus diagnostic criteria [80], presence of spasticity, and independent ambulation (Gross Motor Function Classification System levels I-III).[21] This age range was selected to capture a period of relative gait pattern maturity and gross motor function stabilization in children with cerebral palsy.[22] Following initial screening, 64 individuals were excluded due to the absence of gait videos demonstrating full-body visualization, including the feet. Therefore, a final cohort of 116 participants, contributing 116 gait videos, met all inclusion and exclusion criteria and were included in the subsequent analysis.

### B. Protocol and data collection

To assess the practical applicability of SecurePose in clinical settings, standard gait data collected during clinic visits was used. Participants walked around 15 feet in a straight line along a clinic hallway towards a stationary camera. Videos were recorded using an Apple iPad A1432 at 30 fps and 640x360 pixel resolution. The iPad was handheld by clinic staff, introducing practical challenges commonly encountered in clinical video analysis, such as motion blur, variable focus, lighting changes, and video shaking. SecurePose was specifically developed and validated using this clinical dataset to address these real-world challenges.

### C. Overview of software design

Prior to clinical movement data processing, several essential steps are required, including video pre-processing (video correction), facial anonymization, secure data management and sharing, and accurate extraction of movement kinematics. The logic flow diagram of SecurePose, used to automate these tasks, is presented in Fig. 1. SecurePose is implemented in Python, with a graphical user interface (GUI) developed using TKinter.[23] Hardware and software requirements for SecurePose are detailed in its SourceForge repository.[24] To organize the functions, classes, and variables, we developed two libraries: openHKV (open human kinematics vision) and splib (secure pose library).

D. Project info and meta data

SecurePose initiates with a screen prompting users to enter project information and patient metadata. Users can either load a previously saved project or create a new project by providing a unique name and selecting the videos to be processed, as well as directories for output data. Optional patient metadata can also be associated with the videos. Metadata files can be any information the user wants to link with the videos. These files should share the same base filename as the corresponding video file (e.g., “12345.xlsx” for “12345.mp4”) to enable automatic linking. Upon entering project information, a configuration file and log files are created within a dedicated project directory. This configuration file, created using the Python pickle library,[25] stores all project-related parameters and is loaded upon opening a previously created project. The software supports simultaneous processing of multiple video files for both kinematic extraction and face blurring.

E. Pre-processing: video standardization

Videos recorded from different devices such as mobile phones, tables and cameras can have different hardware biases and video properties. These inconsistencies in file format and video properties could be problematic for body kinematic and face detection algorithms. To solve this problem, SecurePose automatically scans and standardizes all the videos in the project to ensure shared video properties and removed hardware biases. The video standardization done by SecurePose includes correction of video orientation, resolution, file extension, and audio subtraction (i.e. allows the user to mute the video). To perform the video standardization, a custom script, and modules of the Exiftool and OpenCV libraries were used.

F. Body key point extraction

Following video pre-processing, we utilized the Windows implementation of OpenPose to extract the coordinates of 25 body keypoints (Fig. 2), as described in the original OpenPose publication.[19] OpenPose was selected for this work due to its high performance in human pose estimation and inclusion of clinically relevant body keypoints. At each video frame, OpenPose provides two-dimensional (2D) coordinates for each keypoint, along with a confidence level ranging from 0 to 1, representing the algorithm’s certainty regarding the accuracy of the keypoint position. We employed the “BODY 25” model of OpenPose due to its superior accuracy and speed compared to other available models, as well as its inclusion of foot keypoint labels, which are critical for our analysis. To avoid memory errors during keypoint extraction, we utilized the default resolution settings of OpenPose. The extracted body keypoint coordinates were stored separately for each frame in JavaScript Object Notation (JSON) format. The SecurePose GUI allows users to save the OpenPose output videos rendered with extracted body keypoints to a local directory and visualize the kinematics extraction in real-time.

G. Person tracking and person identification

OpenPose extracts body keypoints from videos frame-by-frame, but doesn't track these points across frames. To enable tracking, a custom Python algorithm was developed to identify individuals and assign them unique IDs based on their movement (Fig. 3). The algorithm iterates through video frames (JSON files), tracking the centroid of each OpenPose-detected person. It assigns unique IDs to new individuals while maintaining existing IDs for those already in the frame. Tracking is achieved by calculating the distance between a person's current centroid and their averaged centroid from the previous five frames. The algorithm then updates the JSON files by adding these unique IDs to the extracted keypoints, ensuring each person is uniquely identifiable throughout the video. This approach prioritizes reduced computation time and complexity, relying on the assumption that patient and staff centroids generally won't collide in clinical recordings. While averaging centroids helps estimate position and handle occasional OpenPose errors, a short 5-frame average was chosen to best represent movement given the 30fps video rate.

H. Interpolation

Keypoint coordinates extracted by OpenPose are occasionally erroneous or undetected, particularly for visually occluded body parts. To address this, a custom algorithm was applied to the body keypoints identified for each participant (JSON files; see Fig. 4). Erroneous coordinates were defined a priori as those with an OpenPose confidence level below 0.50. In our dataset, only 0.29% of frames contained erroneous values for at least one body part. Linear interpolation was employed to address these missing coordinates, with separate linear functions fitted for each interval of missing data. This ensured complete keypoint data throughout the video, crucial for reliable face tracking and subsequent blurring. Within the GUI, users can select between interpolating only face coordinates or interpolating the coordinates of the whole body. Interpolating only face coordinates significantly reduces processing time, requiring just 20.8% of the time needed for whole-body interpolation. This option is suitable for users focused solely on facial blurring or who prefer to analyze non-interpolated body kinematics. Conversely, interpolating the entire body's kinematics is more computationally intensive but appropriate for applications requiring linearly interpolated kinematics for further analysis (e.g., [26], [27]). Additionally, SecurePose saves both interpolated and non-interpolated body kinematics of the entire body (including the face) in the output folder, allowing users to utilize raw body coordinates for their specific applications.

I. Patient identification

While tracking all individuals within a video is valuable, for clinical purposes, accurate patient identification is paramount. Several automated approaches are possible, including deep learning, activity-specific movement recognition, and rule-based methods to identify the primary subject. However, deep learning and activity-specific movement recognition require extensive training datasets and may falter in clinically relevant scenarios, such as when patients exhibit minimal movement abnormalities or when non-patient individuals perform similar activities (live walking with the patient). Therefore, we adopted a method identifying the patient based on their prominence within the video recording. Initially, we tracked the movement of the body centroid for all individuals (each assigned a unique ID) throughout the video. Individuals present in fewer than 80% of frames were excluded, assuming clinically relevant videos capture the patient within the camera's field of view for at least 80% of the duration. Subsequently, we computed the average Euclidean distance between each remaining individual's body centroid and the center of the video frame across all frames. The algorithm identifies the individual with the smallest average Euclidean distance to the video frame center (closest to the center) as the primary subject (i.e., patient) in the clinical videos.

J. Blurring

Facial blurring was implemented using a Python script. The script initially reads facial coordinates derived from body key points extracted and stored in JSON files for each individual, separately. The centroid of these coordinates was then calculated using the median values in the X and Y directions to mitigate the impact of potential outliers. Given the potential for high-amplitude head movements in individuals with cerebral palsy, accurate estimation of facial dimensions presents a challenge. Furthermore, the presence of small faces necessitates minimizing estimation errors to ensure effective anonymization. Therefore, facial dimensions were estimated based on the length of the spine – defined as the distance between the neck and pelvis key points – leveraging established human body proportions. Specifically, the facial length and width were assumed to be equal to one-third of the spine length.[28] The GUI provides users with options for person-specific blurring (patient only versus all individuals) and allows selection of the desired blurring type (Gaussian or solid).

K. Quality check

Following automated blurring, the GUI presents an interactive window for quality assessment and refinement. Users can navigate through videos and individual frames to identify potential blurring errors. The GUI enables targeted correction, allowing users to either de-blur specific frames or manually delineate face regions for re-blurring. This interactive process ensures a high level of anonymization quality control.

L. Export data

A window in the menu bar is designed to selectively export the project data to an external directory. Users can export the blurred videos, project back-up files, and coordinates of body key points (as JSON, CSV, or python variables).

M. System specifications

For analysis, a system having the following specifications was used: IntelR Core™ i9 12900 CPU @ 2.40 GHz 16 cores, 32-GB RAM, NVIDIA GeForce RTX3080 GPU with 10GB memory, 64-bit Windows 10 Operating System, and MATLAB 2022b platform. Other hardware and system requirements are mentioned in the software documentation.[24]

N. Effect of face blurring on body key point extraction

To quantify the effect of face blurring on body keypoint extraction accuracy, we compared coordinates extracted from blurred and pre-blurred videos. First, body keypoint coordinates were extracted from the 116 pre-blurred videos using the SecurePose kinematic extraction pipeline. Subsequently, coordinates were extracted from manually face-blurred videos (details of the manual blurring process are described in the following section) using the same SecurePose pipeline. We considered the coordinates extracted from pre-blurred videos as the ground truth, and calculated the error introduced by blurring as the Euclidean distance between corresponding coordinates in pre-blurred and blurred videos. Additionally, we calculated the confidence levels of keypoint estimation for both pre-blurred and blurred videos.

O. Comparison with six existing face blurring techniques

To evaluate the performance of SecurePose, we compared it with six widely used face detection methods using our clinically acquired video dataset. Table I details the trained models employed and the rationale for selecting each method. Since person-specific blurring was not feasible with the six existing methods, SecurePose was configured for “all-person blurring” for this comparative analysis, blurring the faces of all individuals present in the video.

TABLE I
SELECTED EXISTING METHODS FOR THE COMPARISON.

| Model name | Original Article | Download link for trained model | Reason to select |
| --- | --- | --- | --- |
| VJ | [9] | [29] | Fastest and accurate face detector |
| HOG-based | [10] | [30] | Very low computational cost and more accurate than the VJ |
| MMOD | [13] | [31] | Outperformed contemporary methods on the Face Detection Data Set and Benchmark (FDDB) challenge |
| MTCNN | [18] | [32] | Outperformed contemporary methods on FDDB and WIDER Face. But computationally expensive. |
| YOLO-face | [33] | [34] | Outperformed contemporary methods on FDDB and WIDER Face. One of the fastest methods. |
| $S^3FD$ | [14] | [35] | Outperformed contemporary methods on AFW, PASCAL face, FDDB and WIDER Face. |

All methods were compared against a manual face blurring standard: 116 videos of individuals with cerebral palsy walking (dataset described above) were manually blurred by an experienced clinical researcher using Shotcut, an open-source video editing software.[36] The veracity of this manual blurring was independently confirmed by another experienced clinical researcher through frame-by-frame review.

$$IoU = \frac{A_d \cap A_m}{A_d \cap A_m} \tag{1}$$

We assessed the face detection performance of all seven automated methods (SecurePose and the six selected existing methods) against this manual blurring (ground truth) using the 116-video dataset, encompassing a total of 19,753 frames. Intersection over Union (IoU) was calculated for all detected faces and the ground truth using equation (1), where $A_m$ represents the region of the image marked as a face by the manual blurring, and $A_d$ represents the region detected as a face by the algorithm. A detection was considered correct if the IoU value was greater than or equal to 0.5.

Automated methods were evaluated for:

1) True Positives (TP): A correct detection of ground truth face region. (IoU >= 0.5 for ground truth face region)

2) False Positives (FP): An incorrect or a misplaced detection of a face region. (IoU < 0.5 for detected face regions)

3) False Negatives (FN): An undetected face region. (IoU < 0.5 for ground truth face regions)

In the context of face detection, the concept of true negatives (TN) doesn't apply. This is because there is an infinite number of regions representing areas without faces within any given frame [95].

TP, FP, and FN were used to determine:

1) Precision: Precision is the ability of a model, when it detects a face, to have only detected ground truth face regions. A high precision value indicates that the model is able to accurately identify faces in an image without falsely detecting non-faces (see equation (2))

$$Precision = \frac{TP}{TP+FP} = \frac{TP}{All\ detections} \tag{2}$$

2) Recall: Recall is the ability of a model to detect all ground truth face regions. A high recall value indicates that the model is able to detect most of the faces in an image (see equation (3)).

$$Recall = \frac{TP}{TP+FN} = \frac{TP}{All\ ground\ truths} \tag{3}$$

3) F1-score: F1-score can be interpreted as a measure of overall model performance from 0 to 1, where 1 is the best. The F1-score can be interpreted as the model's balanced ability to both capture positive cases (recall) and be accurate with the cases it does capture (precision) (see equation (4)).

$$F1-score = \frac{(2 * Precision * Recall)}{(Precision+Recall)} \tag{4}$$

4) Precision-recall curve: The method used in PASCAL VOC 2012 challenge[37] waws used to calculate precision-recall characteristics. Since both the VJ and the MMOD do not provide confidence levels for their detections, a crucial element for computing precision-recall characteristics, the precision-recall characteristic was not calculated for these methods.

5) Average precision (AP): As described in the PASCAL VOC 2012 challenge[37] 11-point interpolation was used to calculate AP.

P. System Usability Scale (SUS)

To determine the usability of SecurePose in clinical and research settings, ten clinical researchers underwent a training session on how to use the software (~10 minutes), then asked to use the software, and ultimately give feedback on a System Usability Scale (SUS)[38]. SUS scores range from 0 to 100, with a reported average of 68 across published literature. A score exceeding 80.3 indicates software usability within the top 10 percentile.[39] The participating clinical researchers represented diverse professional backgrounds (five with expertise in gait and clinical video analyses, three with experience in clinical video analysis, and two with no prior experience in gait or clinical video analysis), allowing for a comprehensive assessment of software usability across varying levels of expertise.

## RESULTS

A. Effect of face blurring on body key point extraction

To evaluate the effect of blurring on body keypoint coordinate extraction, two metrics were considered (detailed in Section II-N): 1) confidence level (generated for each x, y coordinate pair by OpenPose), and 2) error (Euclidean distance between coordinates for each body keypoint generated by OpenPose pre- and post-face blurring).

Table II presents the mean confidence level for each key body point in pre-blurred and blurred videos across all frames, along with the mean coordinate error extracted from the face-blurred videos. Results indicate a significant decrease in confidence levels for all body keypoints following blurring ($p < 0.05$, one-way ANOVA), with particularly low values (< 0.75) observed for points adjacent to the blurred face (shoulders and chest). Importantly, coordinate differences were observed for all body keypoints between pre- and post-blurred videos, demonstrating that blurring introduced errors in keypoint extraction.

TABLE II
MEAN CONFIDENCE LEVEL FOR COORDINATES EXTRACTION IN PRE-BLURRED AND BLURRED VIDEOS, AND MEAN ABSOLUTE ERROR FOR COORDINATES EXTRACTION IN BLURRED VIDEOS.

| Body keypoint | Confidence level (pre-blurred) | Confidence level (blurred) | Error (pixel) |
|---|---|---|---|
| 0 | 0.868 | 0.315 | 14 |
| 1 | 0.890 | 0.692 | 8 |
| 2 | 0.891 | 0.708 | 7 |
| 3 | 0.920 | 0.905 | 4 |
| 4 | 0.940 | 0.934 | 3 |
| 5 | 0.893 | 0.710 | 8 |
| 6 | 0.908 | 0.895 | 4 |

| 7 | 0.899 | 0.892 | 3 |
|---|---|---|---|
| 8 | 0.948 | 0.934 | 3 |
| 9 | 0.887 | 0.866 | 3 |
| 10 | 0.930 | 0.910 | 4 |
| 11 | 0.866 | 0.844 | 4 |
| 12 | 0.852 | 0.840 | 5 |
| 13 | 0.891 | 0.889 | 4 |
| 14 | 0.886 | 0.872 | 4 |
| 15 | 0.861 | 0.309 | 16 |
| 16 | 0.881 | 0.310 | 15 |
| 17 | 0.847 | 0.250 | 16 |
| 18 | 0.848 | 0.211 | 17 |
| 19 | 0.882 | 0.764 | 4 |
| 20 | 0.859 | 0.803 | 4 |
| 21 | 0.846 | 0.769 | 5 |
| 22 | 0.856 | 0.791 | 5 |
| 23 | 0.849 | 0.834 | 3 |
| 24 | 0.834 | 0.829 | 4 |

While OpenPose is not designed to track individuals across video frames, it should accurately differentiate between individuals and their associated keypoint coordinates within a single frame. However, instances were observed where OpenPose associated keypoint coordinates with the incorrect individual in blurred frames (Fig. 5). Furthermore, blurring led to instances of OpenPose incorrectly identifying left and right body parts (Fig. 5). These findings underscore the importance of performing coordinate extraction before applying face blurring to ensure accurate readings.

## B. Automatic face detection

The performance of each method was evaluated and is presented in Table III, Fig. 6, and Fig. 7. As shown in Table III, SecurePose and three of the six compared methods achieved precision values exceeding 0.9, indicating a high likelihood of correct face detection when a face was identified. The lowest precision was observed with the HOG-based detector (0.689). While some existing methods demonstrated comparable precision to SecurePose, none matched its recall performance. Specifically, the six compared methods exhibited limited ability to detect all ground truth face regions, with recall values ranging from 0.282 to 0.588. Representative examples of blurring errors committed by SecurePose and the other methods are shown in Fig. 8.

SecurePose achieved high precision (0.992) and recall (0.990), yielding a robust F1-score of 0.991, demonstrating its proficiency in accurately and reliably detecting faces within clinical video data. Furthermore, SecurePose exhibited superior precision-recall characteristics and achieved a higher average precision (AP = 0.948) than the other methods assessed (see Table III, Fig. 6 and Fig. 7), suggesting improved face detection performance on clinically recorded gait videos.

Table IV presents a comparison of computational costs. The Viola-Jones algorithm was the fastest method, followed closely by SecurePose. SecurePose demonstrated efficiency by minimizing CPU,

RAM, and GPU resource utilization, though it exhibited slightly higher disk usage compared to other methods. Overall, these results suggest that SecurePose offers a favorable balance between speed and resource consumption relative to the other methods assessed.

TABLE III
PRECESION, RECALL, F1-SCORE AND AVERAGE PRECISION (AP) FOR THE SIX SELECTED EXISTING METHODS AND SECUREPOSE

| Methods | Precision | Recall | F1-score | AP |
|---|---|---|---|---|
| Viola Jones | 0.937 | 0.488 | 0.642 | - |
| HOG-based detector | 0.689 | 0.282 | 0.400 | 0.243 |
| MMOD | 0.822 | 0.376 | 0.516 | - |
| MTCNN | 0.850 | 0.579 | 0.689 | 0.534 |
| YOLO-face | 0.956 | 0.588 | 0.728 | 0.543 |
| $S^3$FD | 0.935 | 0.580 | 0.716 | 0.540 |
| SecurePose | 0.992 | 0.990 | 0.991 | 0.948 |

TABLE IV
COMPUTATIONAL COST OF USING THE METHODS WITH THE SYSTEM SPECIFICATIONS MENTIONED IN THE SECTION II-M.

| Methods | Computational time (minutes) | CPU (%) | RAM (GB) | GPU (%) | GPU memory (GB) | Disk SSD (MB/s) |
|---|---|---|---|---|---|---|
| Viola Jones | 25 | 48 | 0.351 | NA | NA | 0.1 |
| HOG-based detector | 42 | 9 | 0.145 | NA | NA | 0.9 |
| MMOD | 7936 | 10 | 1.942 | NA | NA | 0.1 |
| MTCNN | 153 | 8 | 0.390 | 26 | 8.600 | 0.1 |
| YOLO-face | 66 | 60 | 0.259 | NA | NA | 0.3 |
| $S^3$FD | 334 | 58 | 0.331 | NA | NA | 0.1 |
| SecurePose | 35 | 8 | 0.113 | NA | 0.105 | 2.5 |

### C. Person tracking and identification

While the person tracking and identification strategy implemented in SecurePose likely contributed to higher precision, recall, F1-score, and average precision than existing methods, this methodology had inherent limitations. In 110 of 116 videos (94.8%), the algorithms successfully tracked and assigned unique identifiers (IDs) to all individuals. In 3 videos (2.6%), the unique identifiers of two non-patient individuals were interchanged on 1-2 occasions, attributable to body centroid collisions resulting from close proximity. Algorithm performance was also challenged in these instances by significant occlusion, with the individual in front obscuring >80% of the other's body. In the remaining 3 videos (2.6%), the algorithm failed to assign unique identifiers to non-patient individuals due to extensive occlusion. Importantly, these tracking limitations did not affect the face blurring of any individuals in the videos.

D. Patient identification

The patient identification algorithm operated under the assumption that the individual positioned closest to the frame's center would be the primary subject of the video recording. This assumption yielded a 99% success rate (115 of 116 videos) across the dataset. However, in a single instance, the algorithm incorrectly identified a caregiver accompanying the patient as the primary subject.

This rare error highlights the importance of user awareness during implementation of SecurePose. Specifically, added caution should be exercised when applying patient-specific face blurring to videos containing other individuals positioned centrally within the frame (e.g., an assisting caregiver). In such cases, utilizing the "all person blurring" setting within SecurePose is recommended to ensure complete anonymization.

E. Improvements after quality check

After automated face blurring, a manual quality check was performed. The manual quality checking achieved ceiling performance with an added time of 48 minutes for all 116 videos in aggregate (25 seconds/video). Compared to the time required for ground truth blurring, SecurePose obtained the same performance in less manual effort and time (8.92% of the time taken to blur the videos manually). Please refer to Table V for more details of the performance comparison.

TABLE V
PRECISION, RECALL, F1-SCORE AND TOTAL TIME TAKEN BY SECUREPOSE AUTOMATICS BLURRING AND SECUREPOSE WITH MANUAL QUALITY CHECK ARE SHOWN IN THE TABLE. ALSO, THE TABLE INCLUDES THE TOTAL TIME TAKEN BY GROUND TRUTH MANUAL BLURRING.

| Blurring stages | Precision | Recall | F1- score | Time (minutes) |
|---|---|---|---|---|
| Automated blurring | 0.992 | 0.990 | 0.991 | 35 |
| After manual quality check | 1.000 | 1.000 | 1.000 | 83 |
| Manual Blurring | NA | NA | NA | 930 |

F. System Usability Scale (SUS) results

The average SUS score across these 10 users was 88.75, indicating a high percentile ranking in usability when compared to other software products assessed using the SUS in the literature[39] (Table VI).

TABLE VI
SYSTEM USABILITY SCALE (SUS) SCORE

| **S.No** | **Questionnaire items** | **Average of responses** | **Score contribution** |
|---|---|---|---|
| 1 | I think that I would like to use this software frequently | 3.8 | **2.8** |
| 2 | I found the software unnecessarily complex | 1.9 | **3.1** |

| 3 | I thought the software was easy to use | 4.8 | **3.8** |
|---|---|---|---|
| 4 | I think that I would need the support of a technical person to be able to use this software | 1.2 | **3.8** |
| 5 | I found that various functions in this software were well integrated | 4.8 | **3.8** |
| 6 | I thought there was too much inconsistency in this software | 1.0 | **4.0** |
| 7 | I would imagine that most people would learn to use this software very quickly | 4.8 | **3.8** |
| 8 | I found the software very cumbersome to use | 2.2 | **2.8** |
| 9 | I felt very confident using the software | 4.8 | **3.8** |
| 10 | I need to learn a lot of things before I could get going with this software | 1.2 | **3.8** |
| **Sum** | | | 35.5 |

**Average SUS score = 88.75 (35.5*2.5)**

## DISCUSSION

SecurePose, a novel face blurring algorithm, demonstrates superior performance compared to six existing methodologies when applied to clinically acquired video datasets. Achieving comparable performance to ground truth manual blurring with a significantly reduced processing time, SecurePose facilitates the rapid anonymization and secure sharing of clinically acquired video data.

Quantitative evaluation demonstrates that SecurePose achieves higher precision, recall, F1-score, and average precision for patient face blurring in clinical videos compared to the six existing face detection methods tested. Furthermore, clinical researchers provided a high usability rating for the software. Critically, SecurePose enables the extraction and export of comprehensive full-body kinematic data prior to face blurring, preserving data quality and facilitating subsequent analysis. This automated kinematic extraction capability has the potential to significantly advance clinical assessment and accelerate research in movement disorders.

Person tracking and patient identification algorithms were used to facilitate accurate and automated patient body kinematic extraction. Notably, it achieved a 94.8% success rate in the person tracking and a 99% success rate in the patient identification. The validation results illustrated that SecurePose can perform person-specific face blurring in the vast majority of clinical gait videos. It is additionally notable that this testing occurred on a clinically-relevant data set using videos recorded on a handheld iPad (thus subject to issues with motion, focus, and illumination). The videos of the patients were recorded during a busy clinic (many of whom had masks or clothing covering body key points, thus complicating kinematics extraction and face identification). This real-world data set makes the high performance of SecurePose even more valuable.

A. Limitations and future work

While SecurePose was validated for face blurring and body kinematic extraction on a clinically relevant gait examination, future work should extend validation to other clinically relevant video-based assessments, such as seated or recumbent tasks and other exam maneuvers. Although the primary objective of this study was to develop reliable software for clinical applications, the algorithms were not optimized for computational time. Future iterations will incorporate multi-threading and optimized algorithms to reduce automated kinematics extraction and face blurring processing times. The person tracking, identification, and patient identification algorithms were designed for fast and accurate clinical application; however, performance may vary in non-clinical settings. Future modifications will address adaptability for non-clinical applications. The current patient identification algorithm employs a rule-based approach assuming the patient's presence for >80% of the video duration and proximity to the center. This approach performed successfully in 99% of gait videos; however, alternative rules will be necessary for activities where these assumptions are not valid. Currently, the software operates solely on a local server, ensuring patient data privacy but requiring substantial computational resources. A web-based application leveraging cloud computing would improve accessibility across institutions and facilitate automated kinematics extraction and face blurring.

B. Conclusion

This paper presents the development and validation of open-source software, "SecurePose," for automated face blurring and human kinematic extraction from clinically recorded gait videos. SecurePose achieved a precision of 0.992, recall of 0.990, F1-score of 0.991, and average precision of 0.948 on a dataset of 116 gait videos from individuals with cerebral palsy for automated face detection. Furthermore, the automated process achieved comparable performance to manual blurring in 8.92% of the required time. Usability testing by experienced researchers confirmed the software's user-friendliness, as measured by the System Usability Scale. These findings demonstrate SecurePose's efficacy, efficiency, and usability for both selective and non-selective face blurring and kinematic extraction in clinical gait videos.

FIGURES

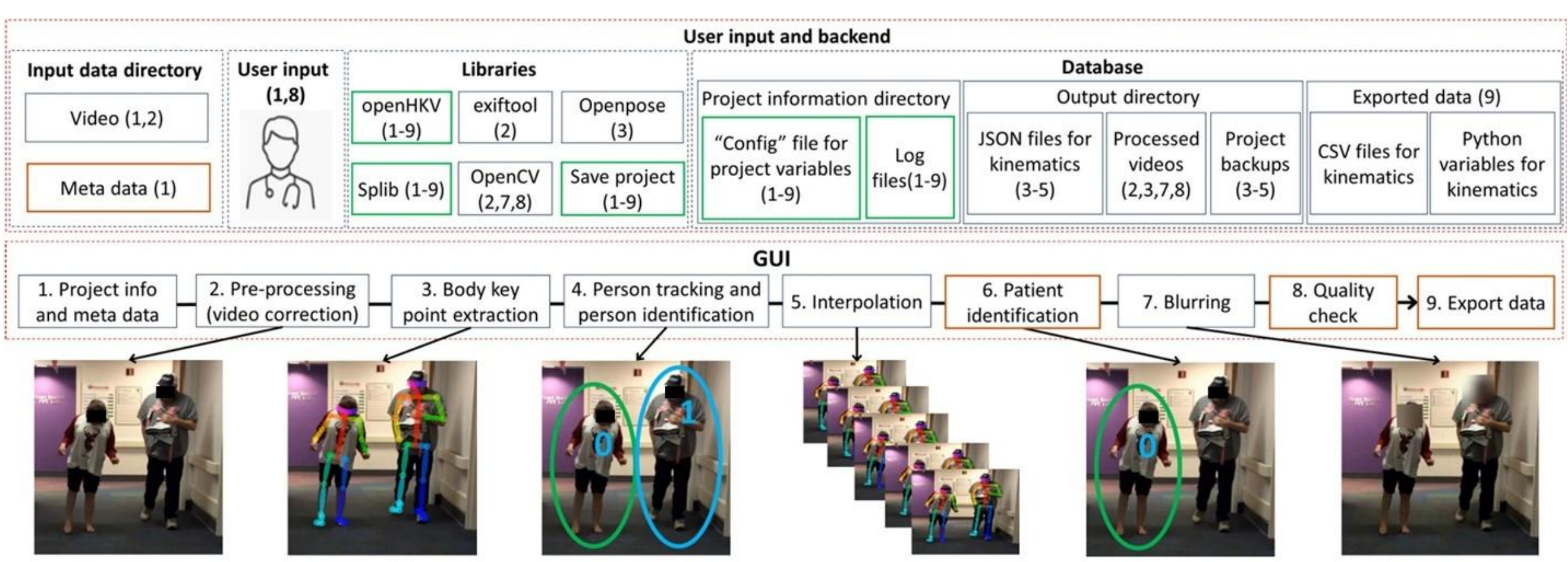

Fig. 1. Logic flow diagram of SecurePose and pictorial representation of some intermediate steps used by SecurePose. The green-colored boundary box modules are called during the execution of each step. The orange-colored boundary box modules are optional steps. Each 'GUI' element represents each of 9 steps taken by SecurePose to perform blurring and kinematics extraction. 'Input and backend' elements are used by SecurePose for input, processing, and output during these 9 steps, as is indicated parenthetically. For example, 'Video' is used during the execution of steps 1 and 2, which are 'Project info and metadata' and 'Pre-processing (video correction)'.

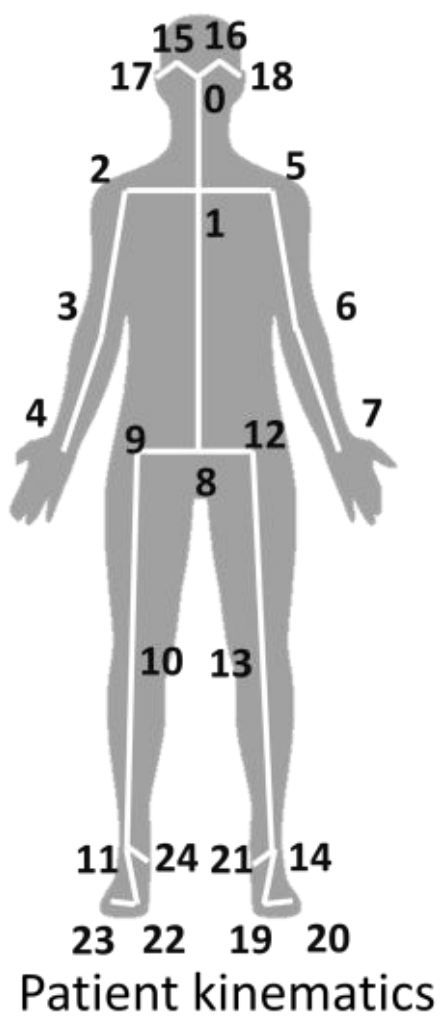

Fig. 2. The index value and location of body key points with respect to the human body.

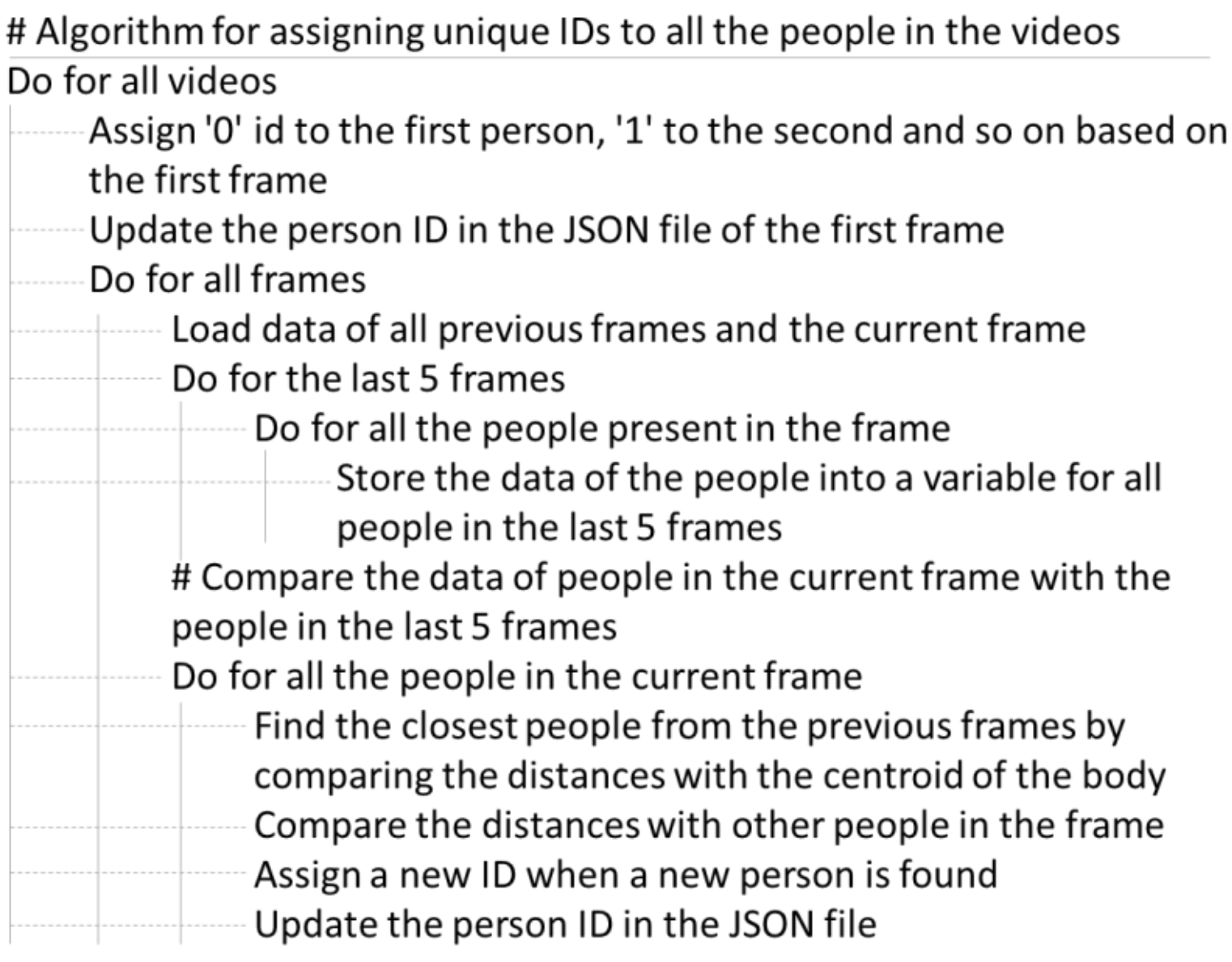

```
# Algorithm for assigning unique IDs to all the people in the videos
Do for all videos
    Assign '0' id to the first person, '1' to the second and so on based on
    the first frame
    Update the person ID in the JSON file of the first frame
    Do for all frames
        Load data of all previous frames and the current frame
        Do for the last 5 frames
            Do for all the people present in the frame
                Store the data of the people into a variable for all
                people in the last 5 frames
        # Compare the data of people in the current frame with the
        people in the last 5 frames
        Do for all the people in the current frame
            Find the closest people from the previous frames by
            comparing the distances with the centroid of the body
            Compare the distances with other people in the frame
            Assign a new ID when a new person is found
            Update the person ID in the JSON file
```

Fig. 3. Algorithm for assigning unique IDs to all the people in the videos of a project.

```
# Algorithm for interpolating the missing coordinates
Do for all videos
    Do for all people in the video
        Do for all the body key points (or only face points)
            Scan for the erroneous and missing frames (bad frames)
            Define a function for linear interpolation for each interval
            of bad frames
            Replace the wrong and missing values with the
            interpolated values
            Update the JSON file of the frames
```

Fig. 4. Algorithm for interpolating the missing coordinates of body key points.

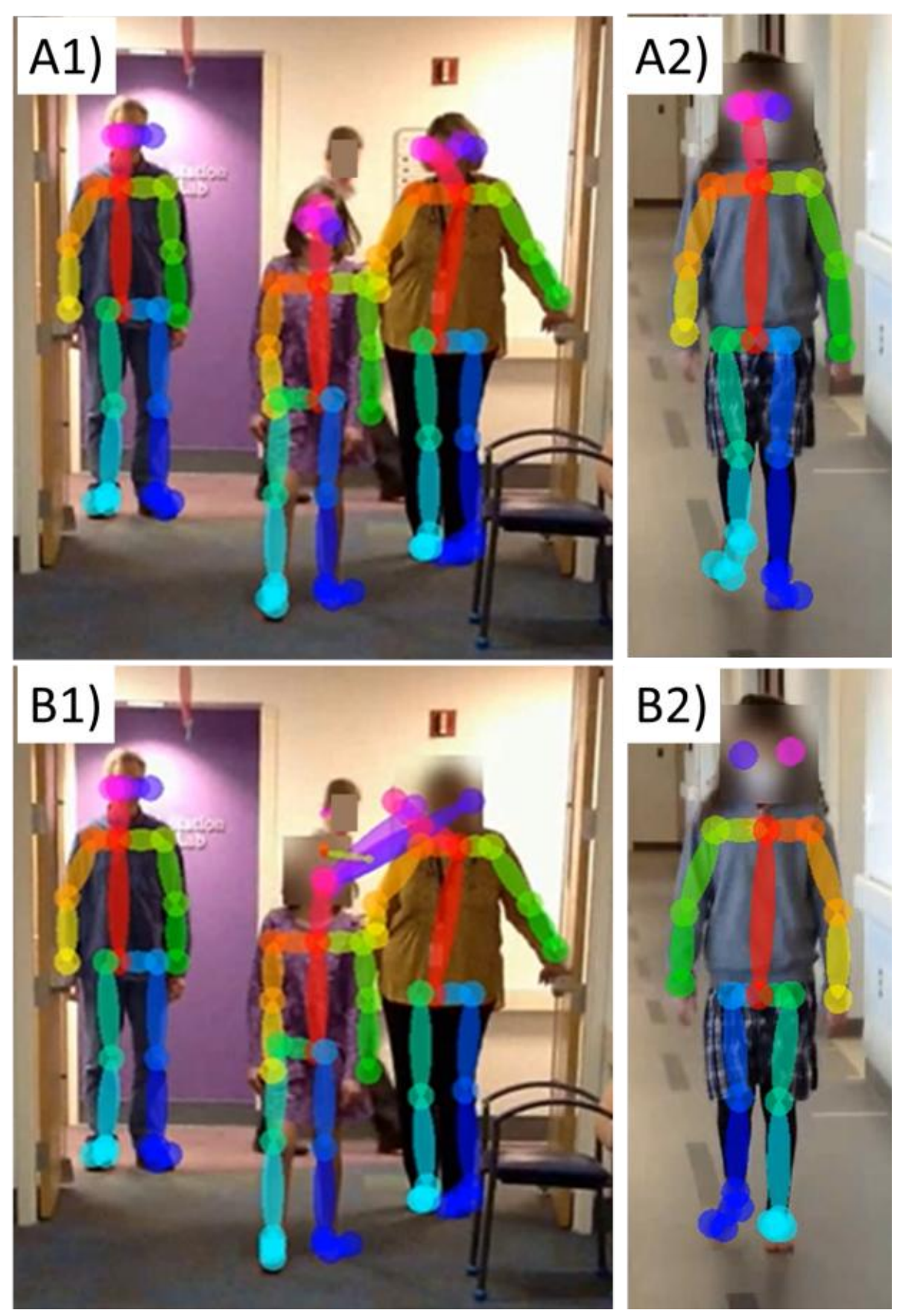

Fig. 5. Examples where kinematic extraction failed in blurred videos. A1 and A2 and the results of kinematic extraction on pre-blurred videos, and B1 and B2 are the results of kinematic extraction on blurred videos. B1) Openpose failed to differentiate between individuals and their associated key point coordinates within the same frame. B2) OpenPose wrongly associated left and right body parts in the blurred video.

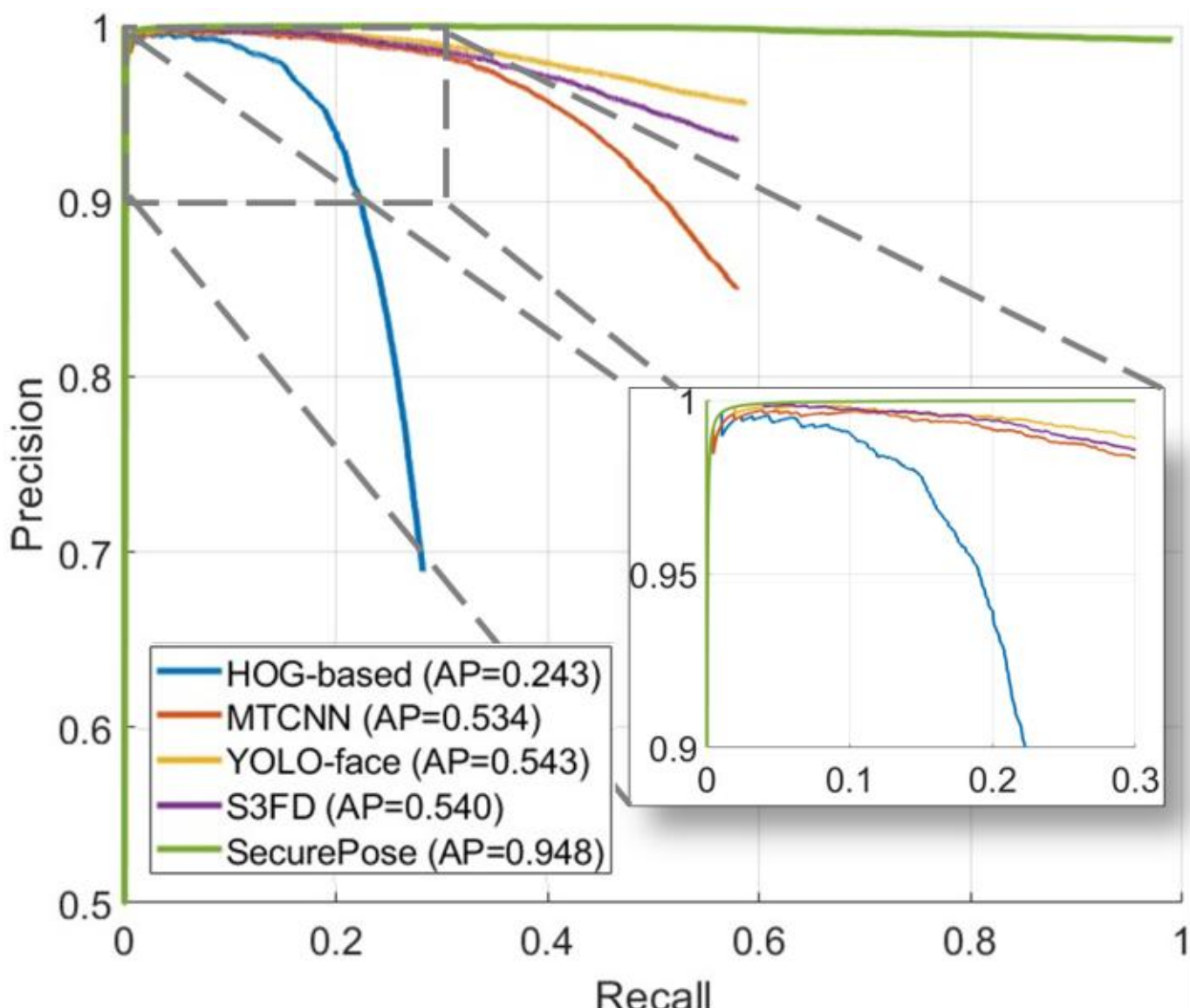

Fig. 6. Precision-recall curve with average precision values for SecurePose and selected existing methods.

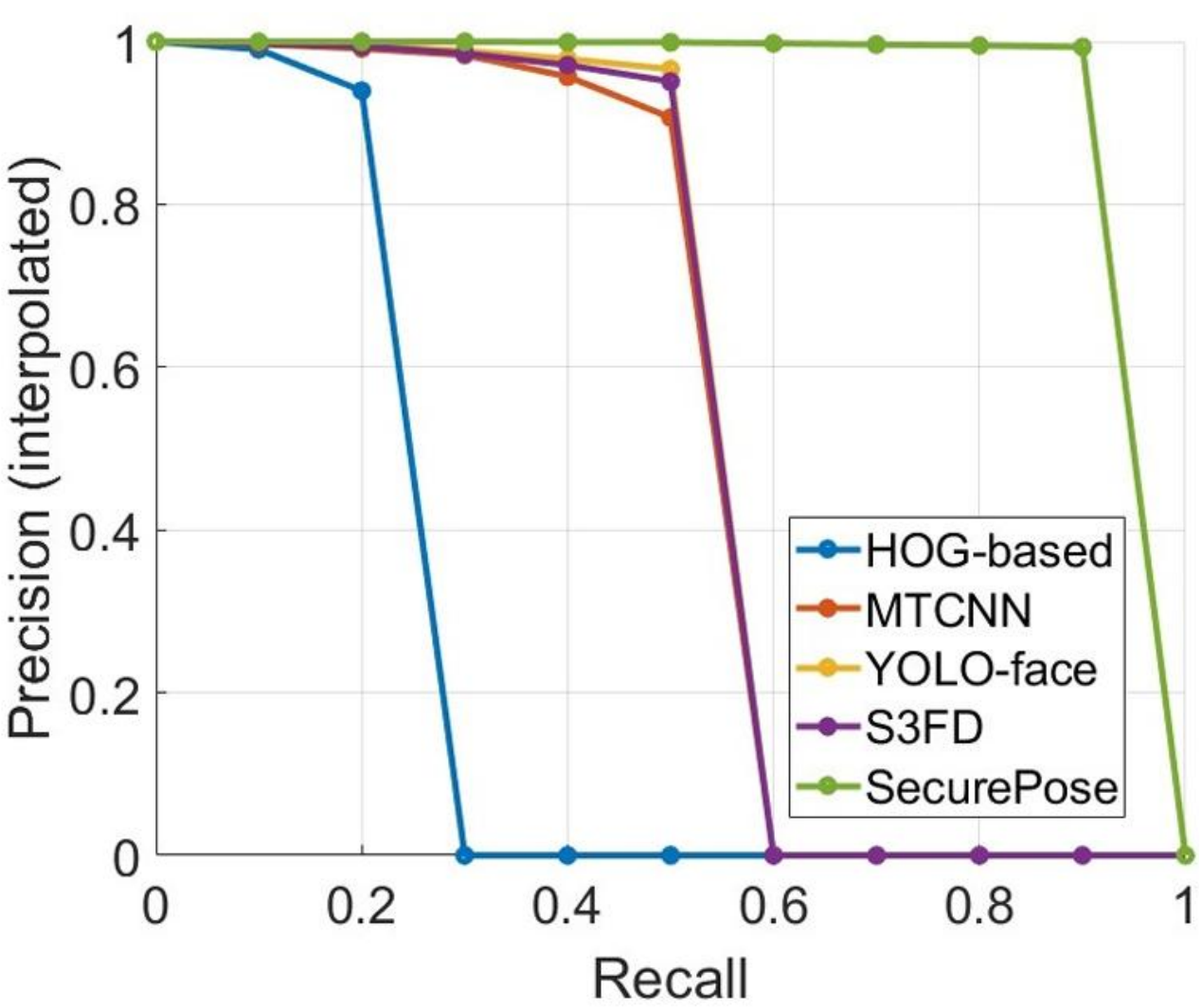

Fig. 7. 11 points interpolated precision-recall curve for SecurePose and selected existing methods.

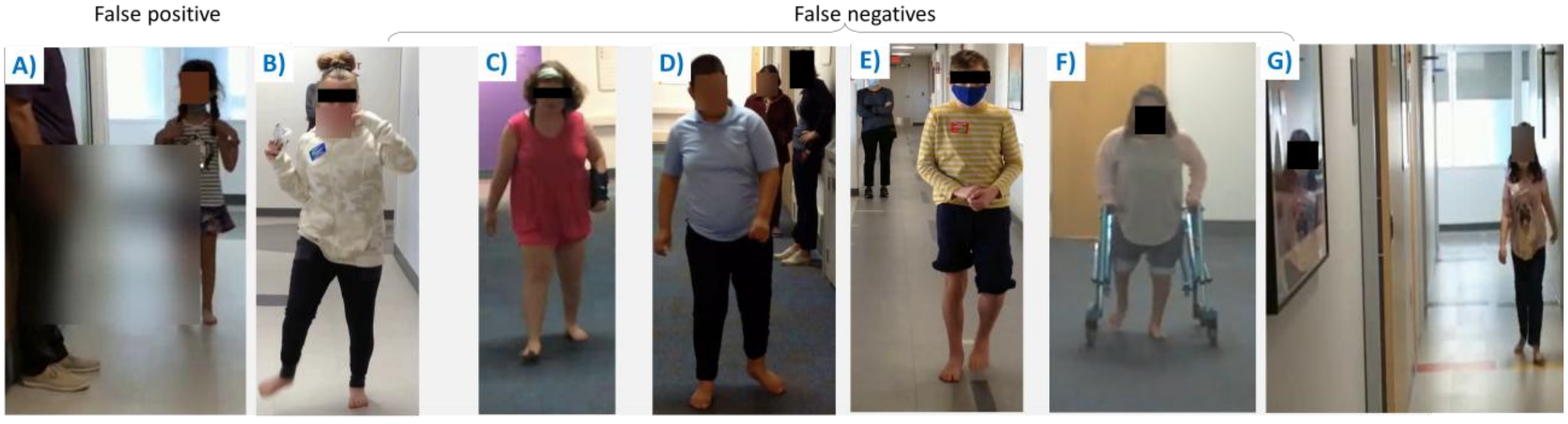

Fig. 8. Examples of some commonly observed failure cases in automated blurring, A) unwanted blurring (Viola Jones), B) half-face unblurred (HOG-based detector), C) full-face unblurred (MTCNN), D) side-face unblurred (ResNet), E) masked-face unblurred (YOLO-face), F) small-face unblurred S3FD and G) unblurred face in reflection (SecurePose). The unblurred face regions are masked with a black patch to hide the identities.